\documentclass[10pt,letterpaper,conference]{IEEEtran}

\usepackage{latex_resources/saurav_macros}
\usepackage{latex_resources/shorthands_minimal}
\usepackage[dvipsnames]{xcolor}
\usepackage{siunitx}
\usepackage[caption=false,font=footnotesize]{subfig}

\pgfplotsset{compat=1.18}

\usepackage{cite}

\newcommand{\normtwo}[1]{\left\lVert#1\right\rVert_2}

\IEEEoverridecommandlockouts                              

\title{\vspace{6mm} \LARGE \bf Scalable Multi Agent Diffusion Policies for Coverage Control}

\author{Frederic Vatnsdal, Romina Garcia Camargo, Saurav Agarwal, Alejandro Ribeiro \thanks{FV, RGC, and AR are with the Department of Electrical and Systems Engineering, University of Pennsylvania, Philadelphia, PA. SA is with the Department of Mechanical Engineering, Indian Institute of Technology Bombay (IITB), India.}%
  \thanks{Emails: \{vatnsdal,rominag,sauravag,aribeiro\}@seas.upenn.edu.}%
\thanks{This work was supported by ARL DCIST CRA W911NF-17-2-0181.}}

\begin{document}
\bstctlcite{IEEEexample:BSTcontrol}
\maketitle

\begin{figure*}[t]
    \centering
    \subfloat{\includegraphics[width=0.53\textwidth, trim={25px 25px 25px 25px}, clip]{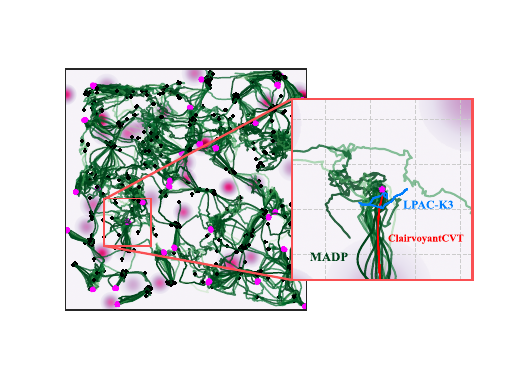}}\hfill
    \subfloat{\includegraphics[width=0.44\textwidth, trim={0px 5px 0px 0px}, clip]{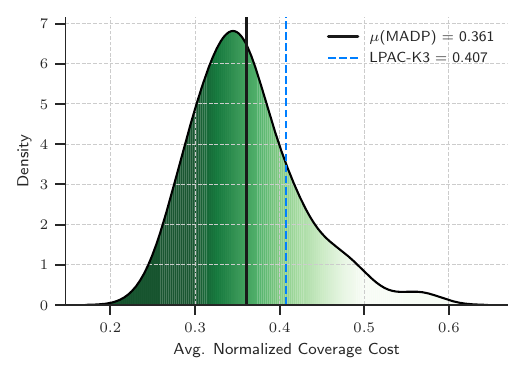}}
    \caption{Demonstration of the Multi Agent Diffusion Policy (MADP) for multi robot collaboration in the coverage control task.
      The inherent stochasticity of MADP allows the generation of a set of finite-horizon trajectories for each robot from their respective positions.
        \textbf{Left:} 
        The coverage control environment with features of interests modeled as purple bivariate Gaussians with randomly sampled means, variances, and positions.
        For each robot, $20$ trajectories (shown in green) are generated corresponding to each run of the MADP for $150$ time steps.
        \textbf{Middle:} The enhanced view captures the trajectories generated by a single robot starting at the magenta dot.
        MADP generates trajectories shown in green (darker shades correspond to a better coverage cost), while the trajectories of LPAC-K3 (blue)~\cite{agarwal_lpac_2025} and Clairvoyant CVT~\cite{cortes2002} (red) are shown for comparison.
        \textbf{Right:} MADP outperforms the state-of-the-art baseline for coverage control on average and occasionally finds solutions with a similar cost to the clairvoyant expert.
        This is especially true when the importance density function (IDF) has much smaller regions of interest than what the models are trained on.%
        \label{fig:moneyshot}}
\end{figure*}

\begin{abstract}
We propose MADP, a novel diffusion-model-based approach for collaboration in decentralized robot swarms. 
MADP leverages diffusion models to generate samples from complex and high-dimensional action distributions that capture the interdependencies between agents' actions.
Each robot conditions policy sampling on a fused representation of its own observations and perceptual embeddings received from peers.
To evaluate this approach, we task a team of holonomic robots piloted by MADP to address coverage control---a canonical multi agent navigation problem.
The policy is trained via imitation learning from a clairvoyant expert on the coverage control problem, with the diffusion process parameterized by a spatial transformer architecture to enable decentralized inference.
We evaluate the system under varying numbers, locations, and variances of importance density functions, capturing the robustness demands of real-world coverage tasks.
Experiments demonstrate that our model inherits valuable properties from diffusion models, generalizing across agent densities and environments, and consistently outperforming state-of-the-art baselines.
\end{abstract}

\section{Introduction}
\label{sec:intro}
In large-scale environments, the use of multi robot teams becomes essential. 
Prior work has proposed strategies for decentralized coordination \cite{agarwal_lpac_2025, longrl, lignn}, yet existing policies often fail to scale or adapt effectively as team size increases. A central difficulty lies in achieving diversity: rather than designing a single navigation strategy, agents must be able to adapt their behavior to the specific demands of the current situation.

Generative Diffusion Models (GDMs) have recently emerged as a powerful framework for high-dimensional tasks~\cite{ho_ddpm_2020_nips}. By capturing complex, multi-modal distributions with stable training dynamics, they are a natural candidate for multi agent control in continuous environments. Recent studies have applied GDMs to trajectory generation in both single and multi agent settings~\cite{chi2024diffusionpolicy, janner_planning_2022, luo2024potentialbaseddiffusionmotion,carvalho_motion_2024, jiang_motiondiffuser_2023, zhu_madiff_2025, shaoul_multirobot_2025, liang2025discreteguideddiffusionscalablesafe, ding2025swarmdiffswarmrobotictrajectory, liang2025simultaneousmultirobotmotionplanning}, demonstrating strong expressivity over short- and long-horizon behaviors, robustness to out-of-distribution scenarios, and promising scalability. However, these works have not yet exploited the full scalability potential of GDMs, with some evaluating for large robot swarms but generating trajectories in a centralized manner.


We propose a collaborative control architecture in which each robot runs a diffusion-policy controller parameterized by a spatial transformer (\scref{sec:diff_models,sec:architecture}).
Following the framework of a learned Perception–Action–Communication (LPAC) loop~\cite{agarwal_lpac_2025}, an agent encodes its local sensor data into a compact feature vector, broadcasts this vector to nearby teammates, fuses incoming messages, and conditions the diffusion process on the combined representation to sample its next control command. The policy is trained centrally by imitating a clairvoyant expert with full state access, but it can be executed in a fully decentralized manner. We evaluate our approach on the coverage control problem~\cite{cortes2002} (\scref{sec:coverage}), where a swarm seeks to minimize a coverage cost defined over an importance density function (IDF). As in~\cite{agarwal_lpac_2025, gosrich2021coveragecontrolmultirobotsystems}, we assume restricted sensing ranges and limited communication radii, which make the problem more challenging while also reflecting realistic operating conditions.

Our model leverages the stochastic and multi-modal properties of diffusion models to learn policies that adapt to diverse coverage demands (\scref{sec:results} and illustrated in \fgref{fig:moneyshot}).
In-distribution evaluations show that our Multi Agent Diffusion Policy (MADP) consistently outperforms state-of-the-art baselines.
Beyond training conditions, MADP demonstrates strong generalization: it adapts to variations in the number, size, and location of importance features, and maintains performance when the number of robots or features is scaled beyond the training configuration. These results highlight both the adaptability and scalability of diffusion-based policies for decentralized multi robot control.
\subsection{Related work}
In robotics, GDMs have been applied to single-agent tasks such as manipulation and motion planning~\cite{chi2024diffusionpolicy, janner_planning_2022}, where actions are generated by iteratively denoising from Gaussian noise.
The denoising process is typically conditioned on observations, obstacle maps, or task goals to guide trajectory generation~\cite{carvalho_motion_2024, luo2024potentialbaseddiffusionmotion}.
Recent work has explored diffusion models for multi agent motion generation.
MotionDiffuser~\cite{jiang_motiondiffuser_2023} and MADiff~\cite{zhu_madiff_2025} use diffusion models with self- and cross-attention for motion prediction and trajectory generation, conditioning on scene context or inter-agent interactions.
Both approaches are intrinsically centralized and do not address or analyze scalability.
The authors in~\cite{shaoul_multirobot_2025} propose a data-efficient framework that incorporates constraints to produce collision-free trajectories, achieving scalability to tens of robots. However, the method remains centralized and emphasizes coordination rather than deeper collaboration between agents. SwarmDiff~\cite{ding2025swarmdiffswarmrobotictrajectory} generates swarm trajectories centrally, which are then refined at the robot level with Model Predictive Control (MPC). Although it employs a diffusion transformer, it does not model agents or spatial structure directly. Liang et al.~\cite{liang2025simultaneousmultirobotmotionplanning} integrate constrained optimization into diffusion sampling, but the combination of nonconvexity and high dimensionality limits scalability. Their follow-up work~\cite{liang2025discreteguideddiffusionscalablesafe} introduces discrete guided diffusion, which decomposes the global planning task into local subproblems, each solved with a diffusion model, improving scalability.

\section{Diffusion Models for Decentralized Multi Robot Policies}
\label{sec:diff_models}
We apply our GDM to a system of $N$ robots with observations $\bfO(t) = [ \bfo_1(t), \cdots , \bfo_N(t)] \in \reals^{d_o\times N}$ and actions $\bfU(t) = [ \bfu_1(t), \cdots , \bfu_N(t)]\in \reals^{d_u \times N}$.
The GDM comprises two processes: forward corruption and backward inference.
The forward process is the iterative corruption of expert actions $\bfU_0$ over $k$ steps with standard white Gaussian noise until convergence to pure noise at $\bfU_K \sim \calN(0, \bfI)$.
This process is described by the Markov chain of conditional Gaussian distributions $q(\bfU_k | \bfU_{k-1})$ and is governed by a sequence of decreasing coefficients $\{\alpha_k\}_{k=1}^K$ called the noise schedule,
\begin{equation}\label{eq:ddpm-fwd-process}
    q(\bfU_k \lvert \bfU_{k-1}) = \calN\left(\sqrt{\alpha_k} \bfU_{k-1}, (1 - \alpha_k)\bfI \right).
\end{equation}
The chain in \eqref{eq:ddpm-fwd-process} can be expressed as the square-root-weighted combination of $\bfU_{k-1}$ and white Gaussian noise $\eps_k \sim \calN(0, \bfI)$,
\begin{equation}\label{eq:fwd-sample}
    \bfU_k(\bfU_{k-1}, \eps_k) = \sqrt{\alpha_k} \bfU_{k-1} + \sqrt{1 - \alpha_k} \eps_k.
\end{equation}
Equations \eqref{eq:ddpm-fwd-process} and \eqref{eq:fwd-sample} require iterating along the chain to obtain every $\bfU_k$, which makes them cumbersome.
In practice, we define $\bar{\alpha}_k := \prod_{j=1}^k \alpha_j$ \cite{ho_ddpm_2020_nips} to obtain $\bfU_k$ directly from $\bfU_0$ and rewrite \eqref{eq:ddpm-fwd-process} as $q(\bfU_k \lvert \bfU_0) = \calN(\sqrt{\bar{\alpha}_k} \bfU_0, (1 - \bar{\alpha}_k)\bfI)$ and \eqref{eq:fwd-sample} as $\bfU_k(\bfU_0, \eps_k) = \sqrt{\bar{\alpha}_k} \bfU_0 + \sqrt{1 - \bar{\alpha}_k} \eps_k$.

GDM inference takes the form of the reverse process in which a fixed prior $\bfU_K \sim \calN(0, \bfI)$ is propelled back down the chain,
\begin{equation}
    p(\bfU_{k-1} \mid \bfU_k, \bfO; \bfH) \sim \calN\left(\mu(\cdot\,; \bfH), \Sigma(\cdot\,; \bfH)\right),
\end{equation}
conditioned on observations $\bfO$ and parameterized by $\bfH$, until $\bfU_0$ is reconstructed.
For expedited sampling with fewer than $K$ backward steps, we use the denoising diffusion implicit model (DDIM) reverse process~\cite{song_denoising_2020_iclr} which can be made deterministic when $\eta = 0, \; (\eta \in [0, 1])$ (see \eqref{eq:ddim-constants}).
Inference invokes the learned noise predictor $\hat\varepsilon(\bfU_k, \bfO; \bfH)$ which we use to define the mean $\mu(\cdot\,;\bfH)$ and we select a fixed covariance $\Sigma(\cdot\;;\bfH) = \sigma_k^2\bfI$, as in \cite{song_denoising_2020_iclr}, 
\begin{equation}\label{eq:ddim-constants}
    \mu(\bfU_k, \bfO; \bfH) = \sqrt{\frac{\bar\alpha_{k-1}}{\bar\alpha_k}}\big(\bfU_k - \frac{1 - \bar\alpha_k / \bar\alpha_{k-1}}{\sqrt{1 - \bar\alpha_k}} \hat\varepsilon\left(\bfU_k, \bfO; \bfH\right) \big)\!.
\end{equation}
To draw a sample $\bfU_{k-1} \sim p\left(\bfU_{k-1} \mid \bfU_k, \bfO; \bfH \right)$, DDIM provides us with a deterministic mapping that preserves the forward marginals $q(\bfU_k \mid \bfU_0)$,
\begin{equation}
       \bfU_{k-1}  = \rho_1 \bfU_k + \rho_2 \hat\varepsilon(\bfU_k, \bfO; \bfH) + \sigma_k \eps_k,
\end{equation}
where $\rho_1 = \sqrt{\frac{\bar\alpha_{k-1}}{\bar\alpha_{k}}}$ and $\rho_2 = \sqrt{1 - \bar\alpha_{k-1} - \sigma_k^2} - \sqrt{\frac{\bar\alpha_{k-1}(1 - \bar\alpha_k)}{\bar\alpha_k}}$ and $\sigma_k = \eta \sqrt{\frac{(1 - \bar\alpha_{k-1})}{(1 - \bar\alpha_k)}} \sqrt{1 - \frac{\bar\alpha_k}{\bar\alpha_{k-1}}}$.

In the context of imitation learning, we are furnished with a dataset of expert actions and observations $\calD := \{\bfU_0^{(m)}, \bfO^{(m)}\}_{m=1}^M$ where $M$ is the total number of training examples.
The inputs to the denoising function are obtained by sampling expert actions and observations from the dataset, $\{\bfU_0, \bfO\}_m \sim \calD$,  the noise schedule step from a uniform distribution $k \sim \calU(1, K)$ and white Gaussian noise, $\eps_k \sim \calN(0, \bfI)$. 
We introduce the forward posterior to demonstrate how we can simplify learning this posterior with $p(\cdot\,;\bfH)$,
\begin{equation}
    q(\bfU_{k-1} \mid \bfU_k, \bfU_0)  = \calN(\Tilde{\mu}(\bfU_k, \bfU_0), \sigma_k^2\bfI)
\end{equation}
where $\Tilde{\mu}(\bfU_k, \bfU_0)$ is the forward posterior mean.
We measure the similarity of distributions $q$ and $p$ with the Kullback-Leibler (KL) divergence. 
Because $q$ and $p$ are both Gaussians with fixed isotropic covariances, the KL divergence is proportional to the squared difference of means under optimization, 
\begin{equation}
    \Ex_q \left[ D_{\text{KL}}\left(q \parallel p\right) \right] \propto \Ex_q \left[ \frac{1}{2\sigma_k^2} \lVert \Tilde{\mu}_k - \mu(\cdot\,; \bfH) \rVert_2^2 \right].
\end{equation}

Expressing the means in terms of the noise $\eps$ that produced $\bfU_k$ enables us to simplify to the standard DDPM loss \cite{ho_ddpm_2020_nips},
\begin{equation}\label{eq:ddpm-loss}
   \calL_{\text{GDM}}( \bfH )  = \Ex_{\bfU_0, \bfO, k, \epsilon} \lVert \epsilon - \hat\varepsilon(\bfU_k, \bfO ; \bfH) \rVert_2^2.
\end{equation}
In this paper, we train a GDM parameterized by spatial transformers (\scref{sec:architecture}) to solve \eqref{eq:ddpm-loss} for a coverage control problem (\scref{sec:coverage}).

\section{The MADP Architecture}
\label{sec:architecture}
\begin{figure}
    \centering
    \vspace{3pt}
    \includegraphics[width=\linewidth, clip]{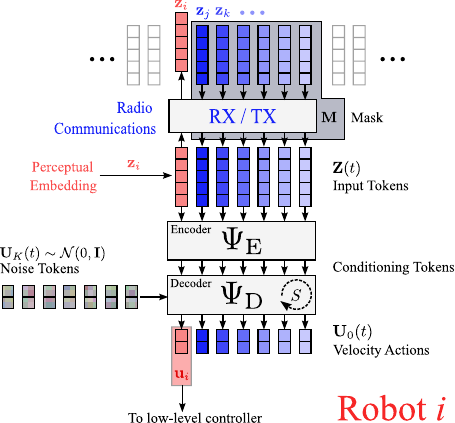}
    \caption{Decentralized sampling of outputs from MADP:
    Each robot in the system runs MADP \textit{locally} and shares perceptual token~$\bfz_i$ with other robots in communication range.
    A token buffer stores the received embeddings from other robots, which are then processed by the encoder $\Psi_{\text{E}}$ once per sample generation.
    The denoiser $\Psi_{\text{D}}$ drives $\bfU_K(t)$ to $\bfU_0(t)$ through $S = 50$ denoising steps (DDIM~\cite{song_denoising_2020_iclr}).
    Finally, the output velocity action $\bfu_i(t)$ is extracted and sent to the low-level controller.%
    \label{fig:block-diagram}}
\end{figure}

We propose a Multi Agent Diffusion Policy (MADP) architecture for the purpose of generating diverse solutions to decentralized and occluded coverage control problems.
Our architecture is constructed as a diffusion model and parameterized by a pair of spatial transformers (ST), a variant of the general transformer architecture in which we impart a rotary encoding scheme \cite{su2023roformerenhancedtransformerrotary} onto the query and key projections.
Spatial transformers have the key characteristic that they can be trained in a centralized environment and deployed in a decentralized communication and computation setting.
The architecture (\fgref{fig:block-diagram}) contains three main components: the Perception Module, the Encoder and the Decoder.
It takes the local maps observed in the environment as input and produces velocity actions for each robot in the system.
We generate environments of size $\SI{1024}{\meter} \times \SI{1024}{\meter}$ for training and evaluation of MADP policies for coverage control.

\subsection{Perception}
Each robot uncovers the occluded portion of the environment using its sensing capabilities defined by a $\SI{64}{\meter} \times \SI{64}{\meter}$ square centered on the robot.
Robots never explicitly share raw information about the environment with each other, i.e., each robot has only its own local view of the environment determined by what it has sensed so far.
We define the \emph{local maps} that each robot processes as the square $(\SI{256}{\meter} \times \SI{256}{\meter})$ centered on the robot ($\SI{1}{\meter}$ per pixel).
Therefore, each robot has access to its historic sensor readings, wider than the sensor view but limited to the local map dimensions.

Conditioning starts with the observation $\bfo_i \in \reals^{4\times 32\times 32}$ made by each robot.
This observation is a set of four local maps downsized from $(\SI{256} \times \SI{256})$ to $(\SI{32} \times \SI{32})$ using bilinear interpolation.
The local maps are: the density function, boundaries and obstacles, neighbors' $x$, and neighbors' $y$.
A three-layer convolutional neural network (CNN) produces an embedding $\bfz_i \in \reals^{32}$ for each observation of each robot.
We collect these embeddings and concatenate the position $\bfx_i$ to produce the input to the encoder $\Psi_{\text{E}}(\bfZ_0; \bfH_{\text{E}})$, where $\bfZ_0 = \left[ \left[ \bfz_1, \bfx_1 \right]^\top, \cdots, \left[\bfz_N, \bfx_N \right]^\top \right] \in \reals^{34\times N}$ and $\bfH_{\text{E}}$ are the parameters of the encoder.

\subsection{Spatial Transformer}
\label{sec:spatialtransformer}
Both the encoder and decoder are realized as spatial transformers~\cite{owerko2025mast}.
A spatial transformer is a set processing architecture composed of attention layers and a relative positional encoding scheme.
The positional encoding must be relative to preserve permutation and shift equivariance, which is necessary because it is impossible to guarantee a fixed permutation during policy inference.
Specifically, we incorporate Rotary Positional Embedding (RoPE)~\cite{su2023roformerenhancedtransformerrotary}.

The dot-product attention mechanism \cite{vaswani_attention_2017} is constructed by projecting the input $\bfZ \in \reals^{d_z \times N}$ into the query space: $\bfQ\bfZ$, into the key space: $\bfK\bfZ$ and into the value space: $\bfV\bfZ$ for $\bfQ, \bfK, \bfV \in \reals^{d_A \times d_z}$, where we have chosen the same dimensions for all spaces.
We compute the attention matrix $\bfA \in \reals^{d_A \times d_A}$ which captures the compatibility  between queries and keys,
\begin{equation} \label{eq:attention_basic}
    \bfA = \textbf{sm}\langle \bfQ\bfZ, \bfK\bfZ \rangle.
\end{equation}
Dot-product attention is finalized by computing the weighted sum of the values, producing the output $\bfY \in \reals^{d_A \times N}$,
\begin{equation} \label{eq:message_passing}
   \bfY = \bfA \bfV \bfZ.
\end{equation}
We combine (\ref{eq:attention_basic}) and (\ref{eq:message_passing}) to express the contextual representation, $\bfY = \textbf{sm}\langle \bfQ\bfZ, \bfK\bfZ \rangle \bfV \bfZ$.

In practice, transformers use multiple layers~$L$ and heads~$H$, i.e., channels.
Then, we can rewrite the self-attention in~\eqref{eq:attention_basic} as a multi-headed layer where the output of the previous layer is the input to the next,
\begin{equation}\label{eq:mhsa}
    \bfY_{\ell}^h = \textbf{sm}\langle \bfQ_{\ell}^h \bfZ_{\ell-1}, \bfK_{\ell}^{h} \bfZ_{\ell-1} \rangle \bfV_{\ell}^h \bfZ_{\ell-1}.
\end{equation}
The outputs of \eqref{eq:mhsa} are subsequently concatenated across heads and passed through the multi-layer perceptron $\bfW_{\ell}$ with a skip connection,
\begin{equation}\label{eq:mhsa_concat}
    \bfZ_{\ell} = \sigma\big(\left[\bfY_{\ell}^1, \dots, \bfY_{\ell}^H\right]\bfW_{\ell} + \bfZ_{\ell-1}\big).
\end{equation}
The learnable parameters at each layer and head are $\bfQ$, $\bfK$, $\bfV$, $\bfW$, where we omit the layer and head notation. 
We implement the non-linearity $\sigma$ as a Leaky ReLU.

To encode positions, we implement RoPE with a sinusoidal basis established by creating a vector of angular frequencies $\bfw = [\omega_1, \cdots, \omega_{D/4}]$ where $D$ is the embedding dimension of the key, query and value spaces.
Pairing adjacent real elements yields $D/2$ complex channels; splitting again between $x$ and $y$ channels yields $D/4$ frequencies per axis.
Frequencies in $\bfw$ are produced by the geometric series $\omega_{i} = 2\pi \tau^{-4i/D}$ where $\tau$ is the encoding period which we select to be proportional to the environment size.
Let $\bfX \in \reals^{2 \times N}$ contain the 2D positions of $N$ agents.
Then, indexing agents with $n$, we can construct the complex matrix $\Tilde{\bfX} \in \complex^{D/2 \times N}$ to house the sinusoids as follows,
\begin{equation}\label{eq:rope_exp}
    [\Tilde{\bfX}]_{i,n} = 
    \begin{cases}
    \exp(j\bfw_{i}[\bfX]_{x,n}) & \text{if $i \leq D/4$} \\
    \exp(j\bfw_{i - D/4}[\bfX]_{y, n}) & \text{if $D/4 < i \leq D/2$}
    \end{cases}
\end{equation}
where $j = \sqrt{-1}$.
We introduce the rotary positional encoding to each head in only the first layer of the transformer,
\begin{equation}
    \bfA_{1}^h = \textbf{sm} \left( \text{Re}\left((\Tilde{\bfX} \odot (\Tilde{\bfQ}_{1}^h \bfZ_{0}))^{\dagger} (\Tilde{\bfX} \odot (\Tilde{\bfK}_{1}^{h} \bfZ_{0}))\right)\right),\\
\end{equation}
where $\Tilde{\bfQ}_1^h, \Tilde{\bfK}_1^h \in \complex^{D/2 \times D}$ are complex renderings of the real-valued query and key matrices, created by pairing adjacent row elements. They are multiplied with the rotary encoding element-wise ($\odot$) and the real part is extracted.
This has the powerful effect of directly embedding relative positions encoded by our Fourier basis into the attention matrix $\bfA$. 
Because the conjugate transpose ($\dagger$) is invoked by the inner product, the complex exponentials defined in (\ref{eq:rope_exp}) produce terms of the kind $\exp(j \bfw_{i} ([\bfX]_{x,n} - [\bfX]_{x,m})) \quad \forall n,m \in [1, N]$, and therefore the positional contribution to $a_{nm}$ depends on the relative difference rather than the absolute positions.

A vital characteristic of the transformer is that it can operate on any number $N$ of received embeddings from other robots.
Therefore, we can train the transformer in a centralized setting and deploy it locally on each robot for decentralized operation.
We have found that adding an attention mask during training aids stability when the density or scale of the scenario changes.
The attention mask is the element-wise \emph{AND} between a windowing mask and a graph component mask $\bfM_{att} = \bfM_{W} \land \bfM_{C}$.
Defining the attention radius in meters as $r_{att}$, we have,
\begin{equation}
\begin{split}
    [\bfM_{W}]_{ij} &= 
    \begin{cases}
    1 & \text{if } \lVert \bfx_i - \bfx_j \rVert_2 < r_{att}\\
    0 & \text{otherwise}
    \end{cases}\\
   \hspace{0.1\linewidth} 
    [\bfM_{C}]_{ij} &= 
    \begin{cases}
    1 & \text{if } \calC(i) = \calC(j) \\
    0 & \text{otherwise}
    \end{cases}
\end{split}
\end{equation}
where $\calC(i) \in \mathbb{N}$ is a mapping from a node to the ID of the graph component the node is currently on.

\subsection{Encoder and Decoder}
The encoder $\Psi_{\text{E}}$ and decoder $\Psi_{\text{D}}$ have similar spatial transformer parameterizations underpinning them.
The primary difference is that the decoder contains cross-attention layers to incorporate the conditioning from the encoder with the input $\bfU_k$ undergoing denoising.
Specifically, each decoder layer is comprised of self-attention followed by cross-attention.
Both components have $L_{ST} = 8$ layers, $H = 8$ heads of dimension $d_{h} = 32$, a geometrically spaced set of encoding frequencies with period $\tau = \SI{1024}{\meter}$  and an attention radius of $\SI{256}{\meter}$.
In the transformers, we use a pre-norm scheme where the layer norm is applied before attention.
During training, we use $T = 1000$ steps of noise over a linearly spaced, decreasing range of $\alpha \in [0.9999, \dots, 0.98]$ and $\alpha_k < \alpha_{k-1}$.
Algorithm \ref{alg:MADP-sampling} details how we can sample from MADP using DDIM.

\begin{algorithm}[t]
  \Input{model $\hat\varepsilon(\cdot\,;\, \bfH)$, observations $\bfO$, DDPM steps $K$, DDIM steps $S$, schedule $\{\alpha_k\}_{k=1}^K$,  noise scale $\eta\in[0,1]$}
  \Output{$\bfU_0$}
  subset $\mathcal{K}=(k_S{=}K>\cdots>k_1 > k_0{=}0)$,
  $\bar{\alpha}_0 \gets 1, \quad \bar\alpha_k \gets \prod_{j=1}^{k}\alpha_j,\quad \forall k=\{1,\ldots,K\}$\\
  $\bfU_{k_S} \sim \mathcal{N}(0,\bfI)$\\
  \textbf{Pre-compute constants and conditioning}\\
  $\sigma_i \gets \eta \sqrt{\frac{1 - \bar\alpha_{k_{i-1}}} {1 - \bar\alpha_{k_i}}} \sqrt{1 - \frac{\bar\alpha_{k_i}}{\bar\alpha_{k_{i-1}}}}, \quad i = 1, \ldots, S$\\
  $c_{0i} \gets \sqrt{\bar\alpha_{k_{i-1}} / \bar\alpha_{k_i}}, \quad i=1, \ldots,S$\\
  $c_{1i} \gets \sqrt{1 - \bar\alpha_{k_i}}, \quad i=1, \ldots, S$\\
  $c_{2i} \gets \sqrt{1 - \bar\alpha_{k_{i-1}} - \sigma_i^2}, \quad i=1, \ldots, S$\\
  $\bfM_{att} \gets \text{compute\_mask(\bfX)}$\\
  $\bfC \gets \Psi_{\text{E}}\left(\text{CNN}(\bfO), \bfX, \bfM_{att}\right)$\\
  \For{$i = S$ \KwDownTo $1$}{
    $\hat\epsilon \gets \Psi_{\text{D}}(\bfU_{k_i},\bfX, \bfC, \bfM_{att})$\\
    sample $\eps \sim \mathcal{N}(0,\bfI)$\\
    $\bfU_{k_{i-1}} \gets c_{0i}\bfU_{k_i} \;+\; (c_{2i} - c_{0i} \, c_{1i})\,\hat\epsilon \;+\; \sigma_i\, \eps$\\
  }
  \caption{MADP Sampling (DDIM)\label{alg:MADP-sampling}}
\end{algorithm}

\section{Decentralized Coverage Amidst Occlusions}
\label{sec:coverage}
Coverage control is the problem of coordinating a swarm of robots to provide effective sensor coverage to areas of variable importance.
We consider the coverage problem on a planar environment, $\calX \subset \reals^2$.
A scalar field called the \emph{importance density function} (IDF) maps points in the environment to values of importance, $\Phi \colon \calX \to \reals_+$.
The IDF is occluded and therefore unknown to each robot at $t_0$, save for the small sensing region surrounding each robot.
Robots are modeled with single integrator dynamics.
The robot actions are bounded in norm, such that $\normtwo{\bfu_i(t)} \leq u_{\max}$.
The system's dynamics are then,
\begin{equation}
   \bfX(t + \Delta t) = \bfX(t) + \Delta t \bfU(t).
\end{equation}

The coverage cost is measured by accumulating the distance between the closest agent and a given point $\bfv$ in the environment, weighted by the IDF at that point, 
\begin{equation} \label{eq:coverage_cost}
    \calJ(\bfX(t)) = \int_{\bfv \in \calX} \min_{i\in \{1,\ldots,N\}} f(\lVert \bfx_i(t) - \bfv \rVert_2) \Phi(\bfv) d\bfv.
\end{equation}
Computing the integral in~\eqref{eq:coverage_cost} is impractical due to the minimum that needs to be found for each point.
Instead, we use Voronoi tessellation, which decomposes the environment into cells~$V_i$, each assigned to a robot~$i$, such that the robot~$i$ is closest to any point in its corresponding cell~$V_i$.
The positions of the robots induce the Voronoi tessellation $\calT := \{V_i\}_{1:N}$ of $\calX$ and the cells emerge from,
\begin{equation} \label{eq:voronoi_partitioning}
    V_i = \{ \bfv \in \calX \: \mid \: \lVert \bfx_i - \bfv \rVert_2 \leq \lVert \bfx_j - \bfv \rVert_2, \forall \; j \neq i\}.
\end{equation}
The objective function~\eqref{eq:coverage_cost} is made computationally tractable by the tessellation $\calT$ and we also select $f(x) = x^2$,
\begin{equation} \label{eq:coverage_cost_voronoi}
    \calJ(\bfX(t)) = \sum_{i=1}^N \int_{\bfv \in V_i} \lVert \bfx_i(t) - \bfv \rVert_2^2 \Phi(\bfv) d\bfv.
\end{equation}


For our experimental setup, we consider holonomic robots that have limited observability, i.e., limited knowledge about the state of the environment $\calX$ and its discovery, and the position of other robots $\bfX(t)$.
The sensor field of view (FOV) is defined by the square $r_s \times r_s$ centered around each robot.
A communication radius $r_{c}$ is defined to determine which other robots in the swarm are within robot $i$'s neighborhood.
The nature of our architecture, however, allows attention to be applied between agents that are more than one hop away.
This introduces a new notion of neighborhood as agents within the attention mask defined in \scref{sec:spatialtransformer}.
We denote the set of agents within (masked) attention of agent $i$, as $\calM_i$.
Robot~$i$ will have access to information obtained from the local environment, as well as positions~$\bfx_j$ and embeddings~$\bfz_j$ from neighboring robots, $j \in \calM_i$. 
Our setting has ideal wireless conditions with neither delays nor errors in transmission.

\subsection{Imitation Learning}
We create a dataset $\calD = \{\bfU_0^{(m)}, \bfX^{(m)}, \bfO^{(m)}\}$ of $100\times10^3$ examples by sampling observations and actions from rollouts of the clairvoyant Centroidal Voronoi Tessellation (CVT) expert (see \hyperref[app:baselines]{Appendix}).
The dataset is generated in environments of size \qtyproduct{1024x1024}{\metre} containing $N = 32$ robots and $F = 32$ Gaussian features.
The coverage control simulator and data generation code are taken from the Coverage Control Library~\cite{agarwal_lpac_2025}.
We divide this dataset into \num{70000} training examples, \num{20000} validation examples and \num{10000} test examples.
The learning hyperparameters, including the architecture parameters, were selected by performing a Bayesian sweep over the learning rate and weight decay.
MADP was trained with a learning rate of $\text{lr} = \num{8.5e-5}$ and a weight decay of $\text{wd} = \num{2.1e-12}$.
The model was trained for $1000$ epochs and a batch size of $196$ on an Nvidia RTX 3090 GPU; the validation loss patience was set to $500$ with a minimum error of $\num{1e-4}$.

\section{Results}
\label{sec:results}
\begin{figure}[htpb]
    \centering
    \includegraphics[width=\linewidth]{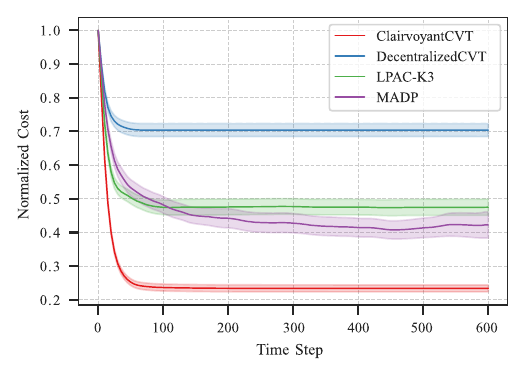}
    \caption{
    Normalized coverage cost as each policy is rolled out over $50$ randomly generated environments with $N=32$ robots, $F = 32$ Gaussian features in the IDF over $600$ time steps.
    The dark line of each trace is the mean cost, while the band shows the $95\%$ confidence interval.
    Lower values indicate better performance.
    Both LPAC-K3~\cite{agarwal_lpac_2025} and the proposed MADP perform substantially better than the baseline Decentralized CVT algorithm, with MADP performing better than LPAC-K3.%
    \label{fig:coverage-basic}}
\end{figure}

\begin{figure}[htpb]
    \centering
    \includegraphics[width=\linewidth]{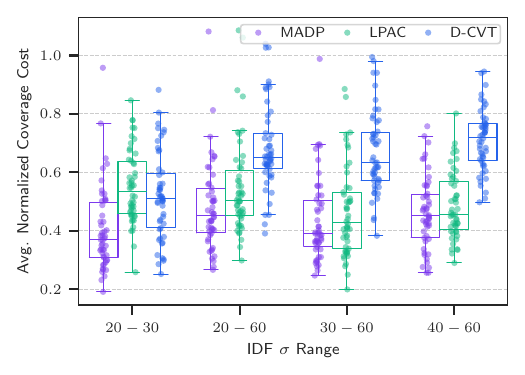}
    \caption{Normalized cost distribution obtained from evaluating MADP, LPAC-K3, and DCVT on 50 unseen environments for varying standard deviations ($\sigma$) of importance of IDF features.
    For training, the $\sigma$ for each feature was randomly sampled from $[40,60]$, and the policies are tested on different ranges of $\sigma$, i.e., out-of-distribution IDFs.
    MADP consistently performs better than the baselines and demonstrates greater adaptability when the size of the Gaussian features is aggressively constrained.\label{fig:sigma-change}}
\end{figure}

\begin{figure*}[htbp]
    \centering
    \hspace{0.1cm}
    \subfloat{\includegraphics[width=0.23\textwidth, trim={70px 59px 29px 40px}, clip]{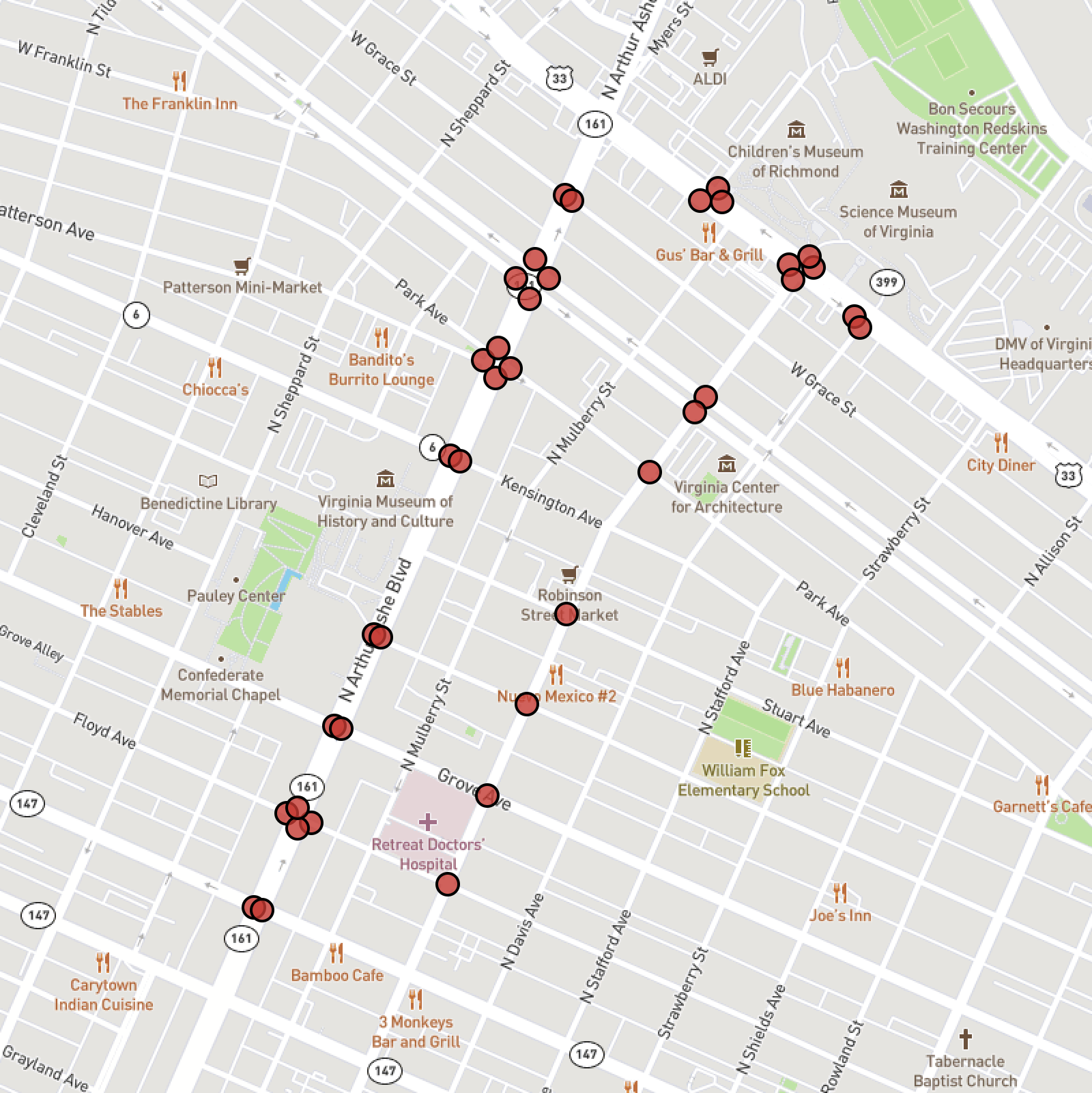}}\hfill
    \subfloat{\includegraphics[width=0.23\textwidth, trim={70px 59px 29px 40px}, clip]{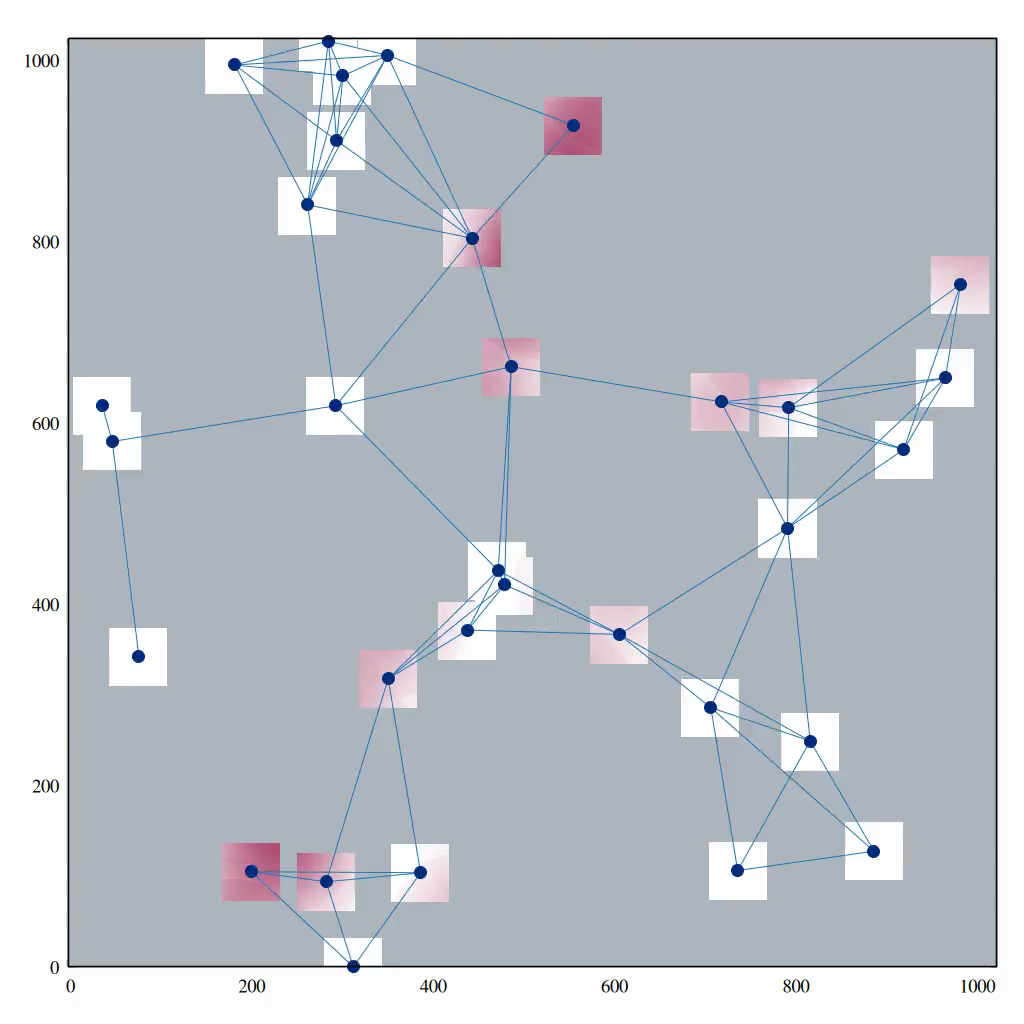}}\hfill
    \subfloat{\includegraphics[width=0.23\textwidth, trim={70px 59px 29px 40px}, clip]{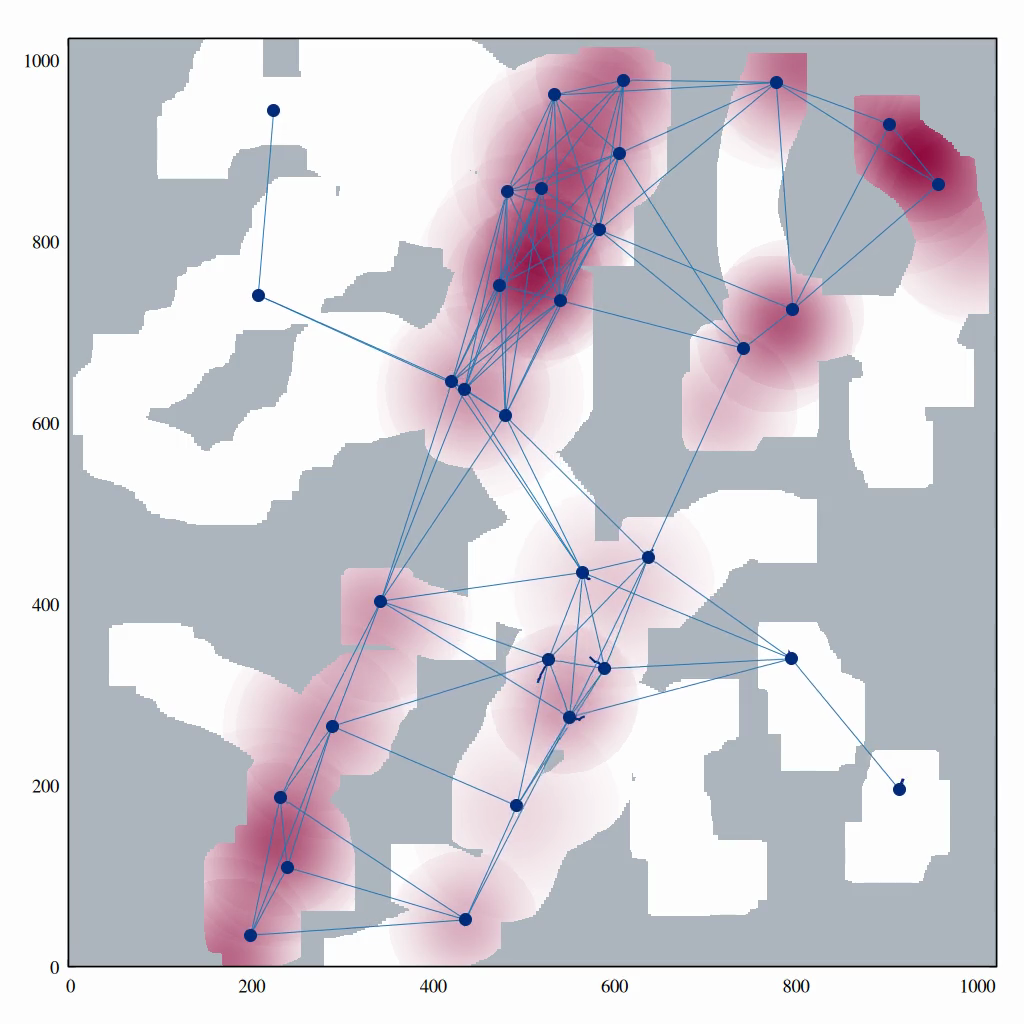}}\hfill
    \subfloat{\includegraphics[width=0.23\textwidth, trim={70px 59px 29px 40px}, clip]{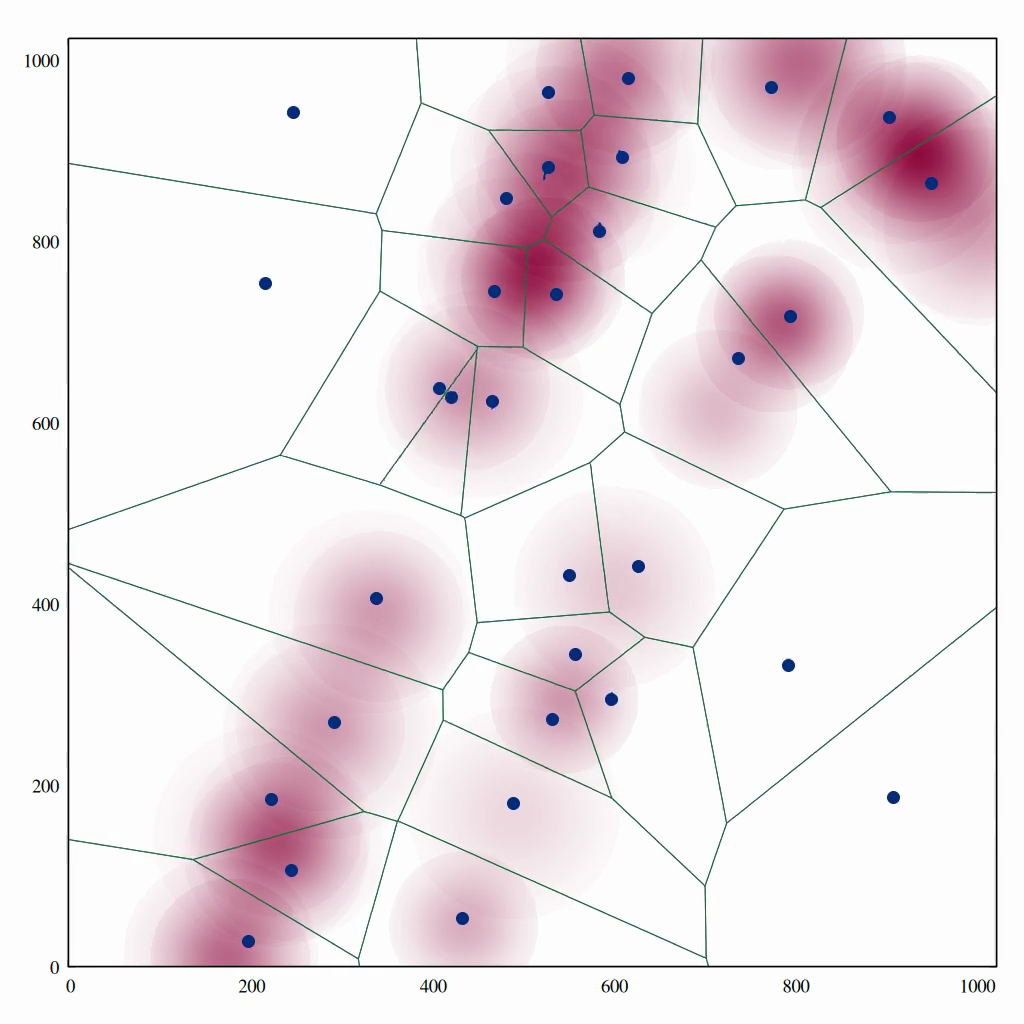}}\hspace{0.1cm}
    \\
    \includegraphics[width=\textwidth, trim={0px 0px 10px 0px}, clip]{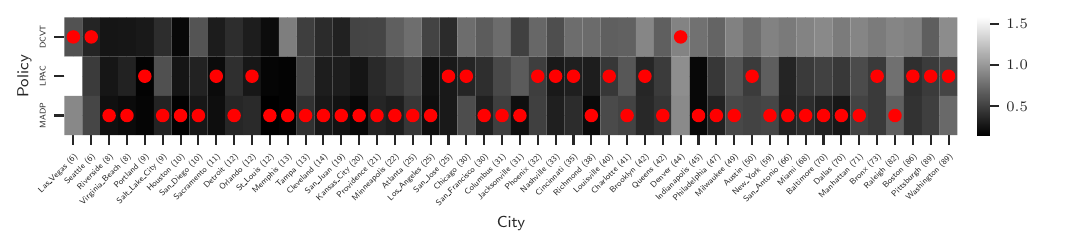}
    \vspace{-0.75cm}
    \caption{
        \textbf{Top:}
        We use the locations of traffic lights from a city (left) to inform the generation of the importance density function (fully visible; right).
        Shown here: a rollout of MADP on the Richmond map, running the policy over 600 timesteps.
        From the second image on the left to right, the timesteps are $0$, $300$, and $600$.
        In the last frame, we reveal the complete IDF and overlay the Voronoi tessellation \eqref{eq:voronoi_partitioning}.
        \textbf{Bottom:}
        The heatmap shows the normalized coverage cost of policy rollouts over 600 time steps for each city.
        Darker values indicate better performance.
        A red dot is placed on the cell of the best-performing policy for each city.
        The initial positions of the $N = 32$ robots are sampled from a uniform distribution in the environment.
        The parenthetical next to the city name indicates the number of Gaussian Features $F$.
        The scenario configuration is the same across controllers.
        Map data is from OpenStreetMap \cite{OpenStreetMap_contributors}.
    }
    \label{fig:coverage-real-world}
\end{figure*}

We present our simulated experimental results to quantify MADP's capabilities.
Unless otherwise stated, we consider an environment $\calX$ of size $(\SI{1024}{\meter} \times \SI{1024}{\meter})$.
Each scenario has $N = 32$ holonomic robots, each with a communication radius of $r_c = \SI{256}{\meter}$, and sensor aperture of $(\SI{64}{\meter}\times \SI{64}{\meter})$.
A total of $F = 32$ bivariate Gaussian features $\phi_i = \calN(\bfmu_i, \sigma_i^2\bfI) \quad \forall i \in \{1, \ldots,F\}$ are placed randomly in the environment, such that $\bfmu_i^{(x)}, \bfmu_i^{(y)} \sim \calU(0, 1024)$, with variable variance, $\sigma_i \sim \calU(40, 60)$.
We restrict our Gaussian features so that their peak density values lie in the range $[0.6, 1.0]$ and the distribution is truncated at $2\sigma_i$. 
For an example of an environment, see \fgref{fig:coverage-real-world}.
The primary evaluation metric is the coverage cost defined in (\ref{eq:coverage_cost_voronoi}), normalized by the initial coverage cost at $t = 0$. For a description of the baselines, see \hyperref[app:baselines]{Appendix}. 
For $N=32$, MADP inference on an RTX 3090 with FP32 is $0.36 \pm \SI{0.006}{\second}$ per step.

\fgref{fig:coverage-basic} shows the normalized coverage cost over $600$ rollout steps, averaged across 50 unseen, randomly generated environments.
The model was trained with $N=32$ robots and $F=32$ Gaussian features.
MADP significantly outperforms DCVT.
MADP also outperforms LPAC after about 100 steps.
This experiment serves as an in-distribution evaluation of the proposed policy.
Subsequent subsections examine out-of-distribution settings with varying numbers of robots, Gaussian features, and environment configurations.

\subsection{Adaptability to Small Areas of Interest}
We evaluate MADP in environments where the Gaussian features are reduced in size, controlled by adjusting the range of values for $\sigma_i$.
This scenario is relevant for aerial robots, where altitude changes can alter the scale of observed Gaussian features (i.e., features appear smaller at higher elevation).
Results are shown in \fgref{fig:sigma-change} for different ranges of $\sigma$.
Only the range $\sigma \in [40, 60]$ is in-distribution for both MADP and LPAC; the remaining ranges contain sizes of features not seen during training.
MADP achieves the lowest normalized coverage costs, demonstrating effective generalization to environments with varying distributions.
Performance remains consistently strong across all tested values, with only a few outliers.

\subsection{Real-World Scenarios}
\begin{figure*}[htbp]
    \centering
    \subfloat{\includegraphics[width=0.23\textwidth, trim={70px 59px 29px 40px}, clip]{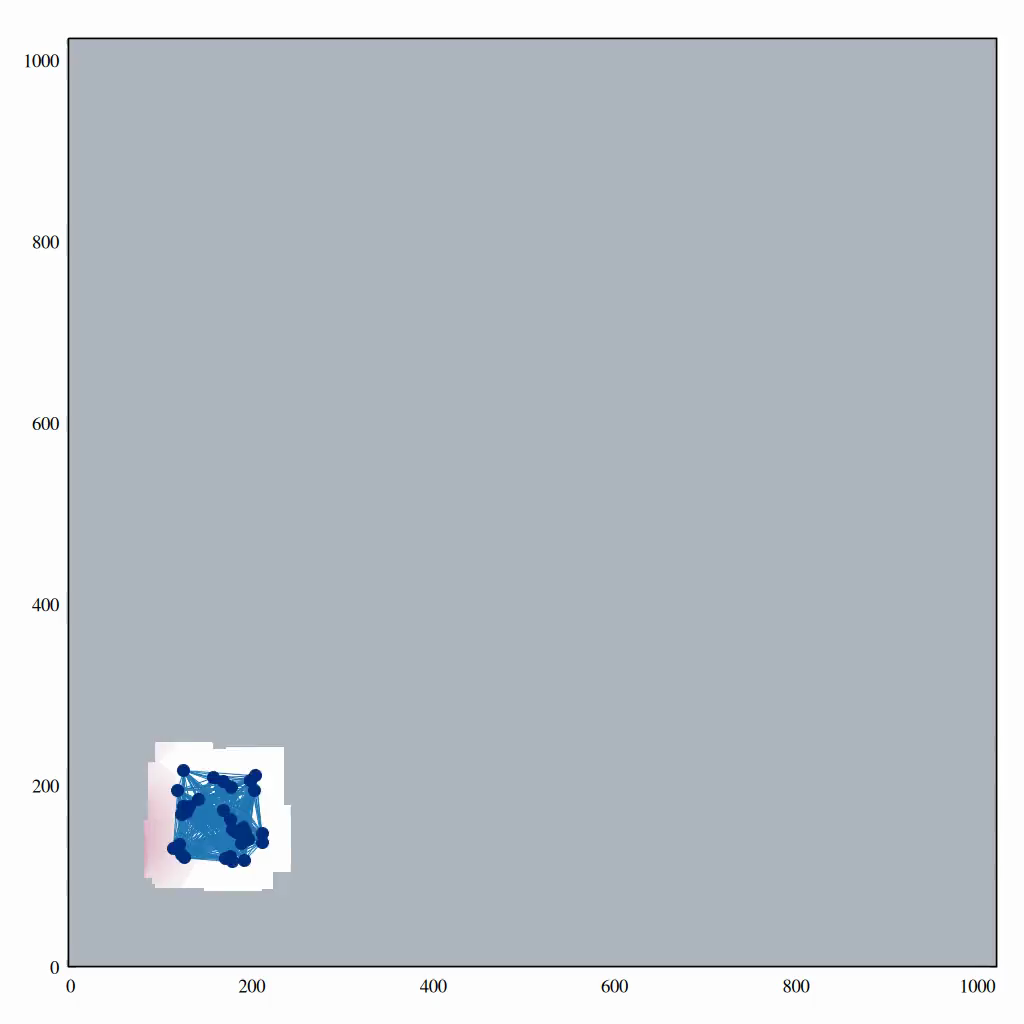}}\hspace{0.1cm}
    \subfloat{\includegraphics[width=0.23\textwidth, trim={70px 59px 29px 40px}, clip]{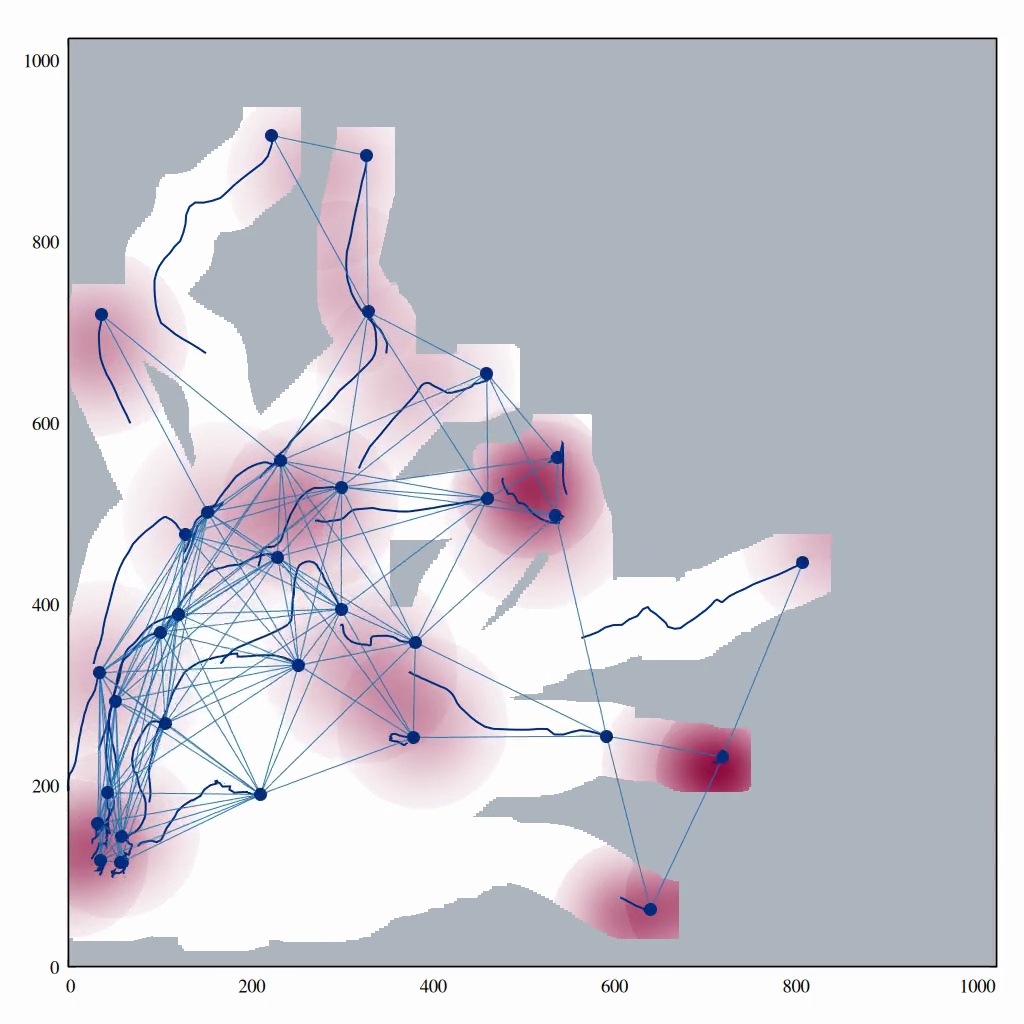}}\hspace{0.1cm}
    \subfloat{\includegraphics[width=0.23\textwidth, trim={70px 59px 29px 40px}, clip]{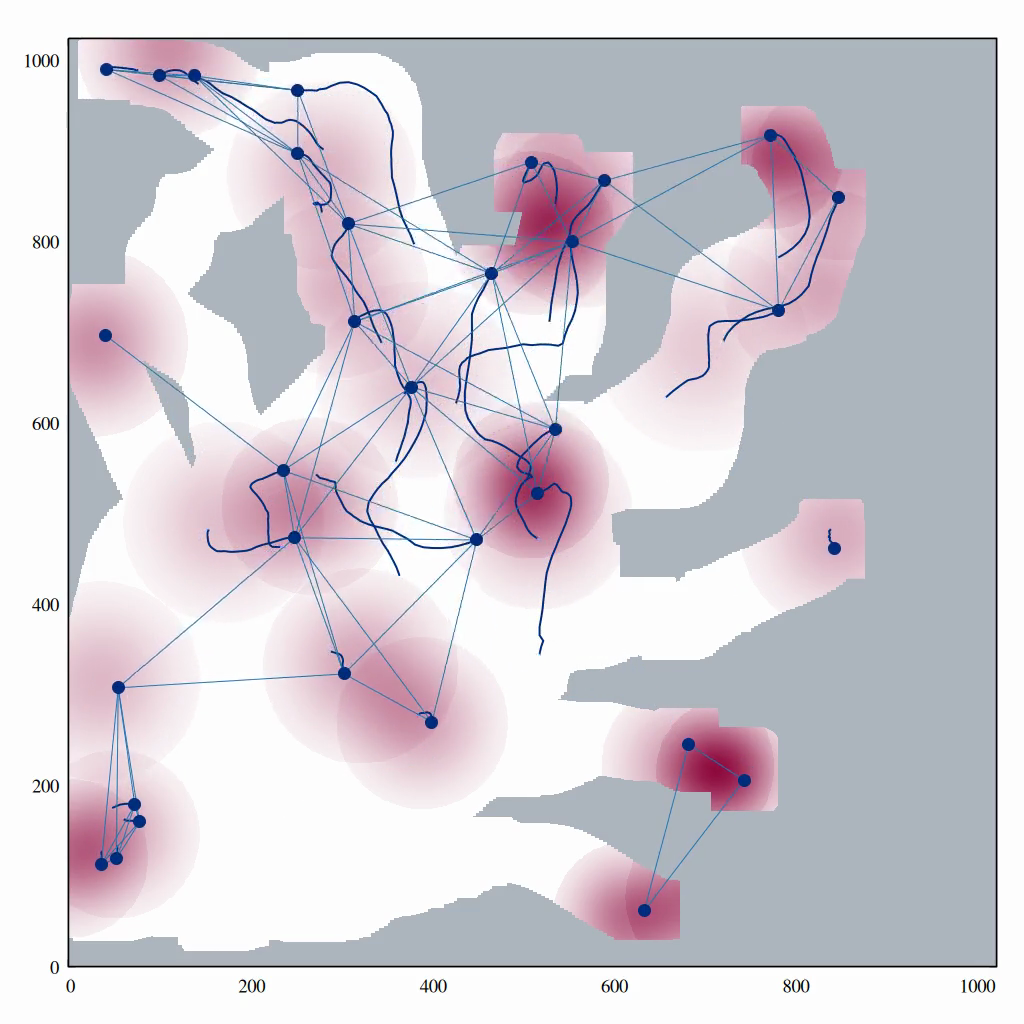}}\hspace{0.1cm}
    \subfloat{\includegraphics[width=0.23\textwidth, trim={70px 59px 29px 40px}, clip]{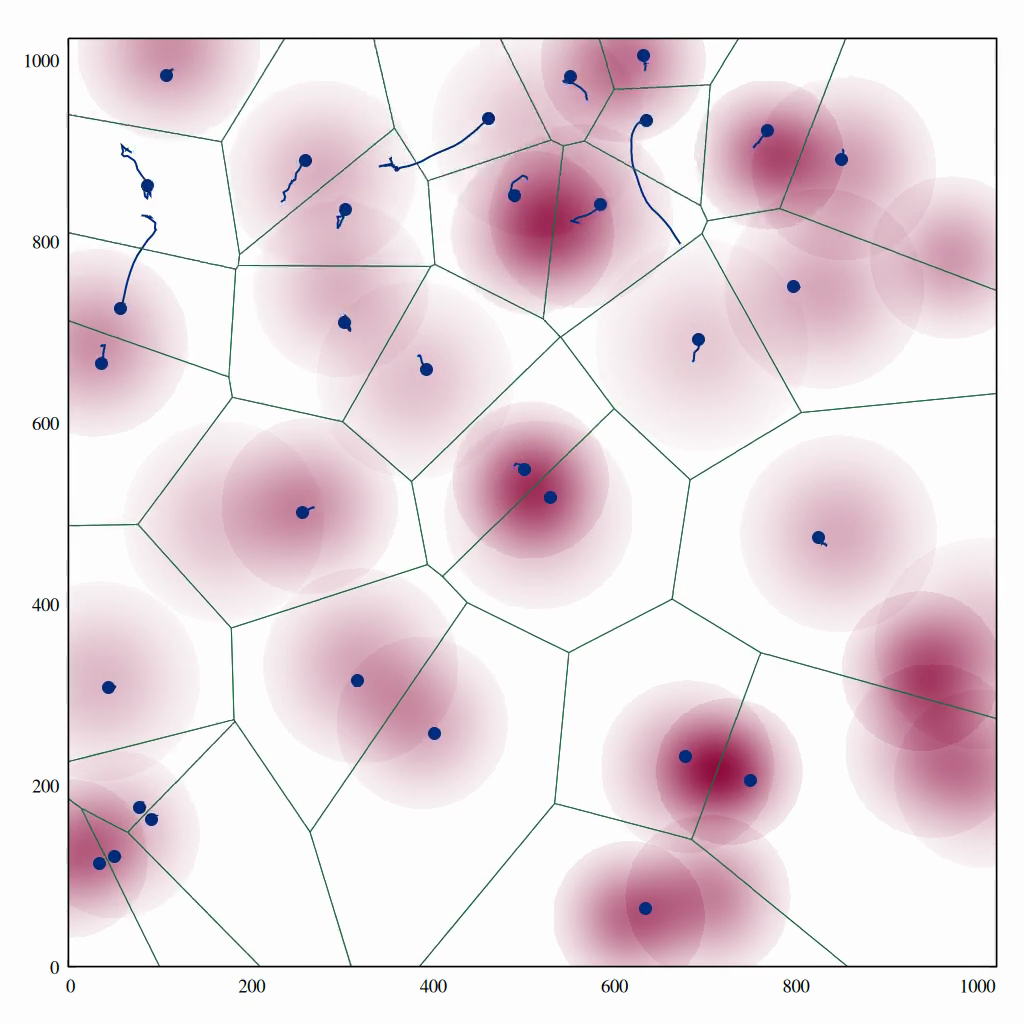}}\hspace{0.1cm}
    \\
    \subfloat{\includegraphics[width=0.23\textwidth, trim={70px 59px 29px 40px}, clip]{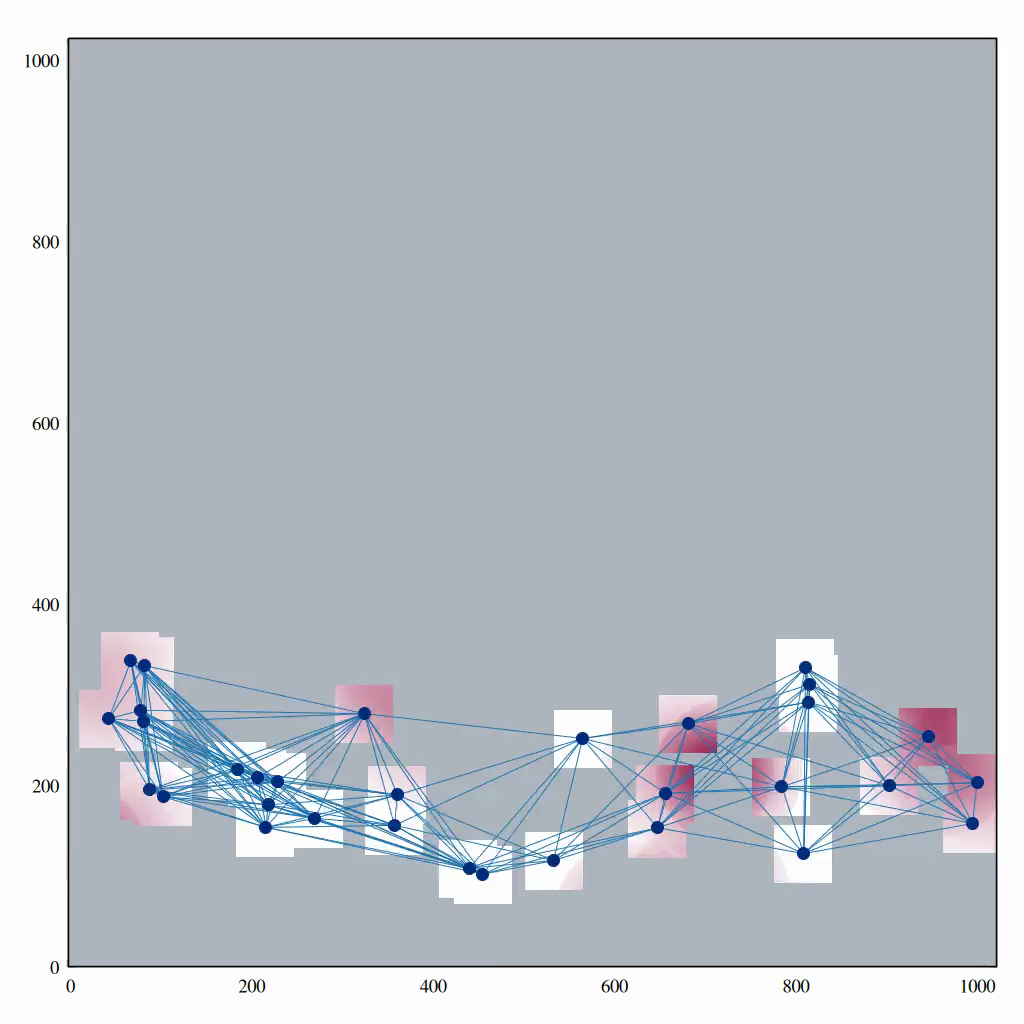}}\hspace{0.1cm}
    \subfloat{\includegraphics[width=0.23\textwidth, trim={70px 59px 29px 40px}, clip]{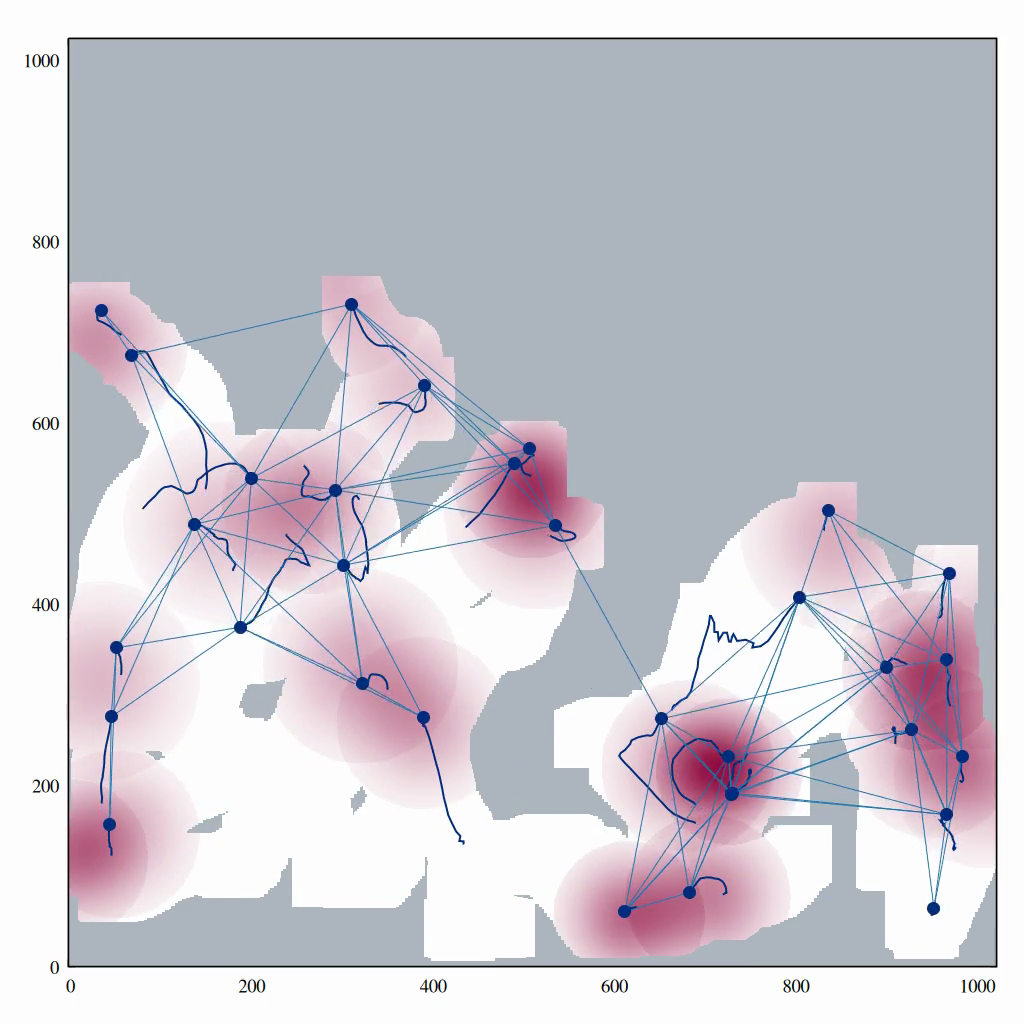}}\hspace{0.1cm}
    \subfloat{\includegraphics[width=0.23\textwidth, trim={70px 59px 29px 40px}, clip]{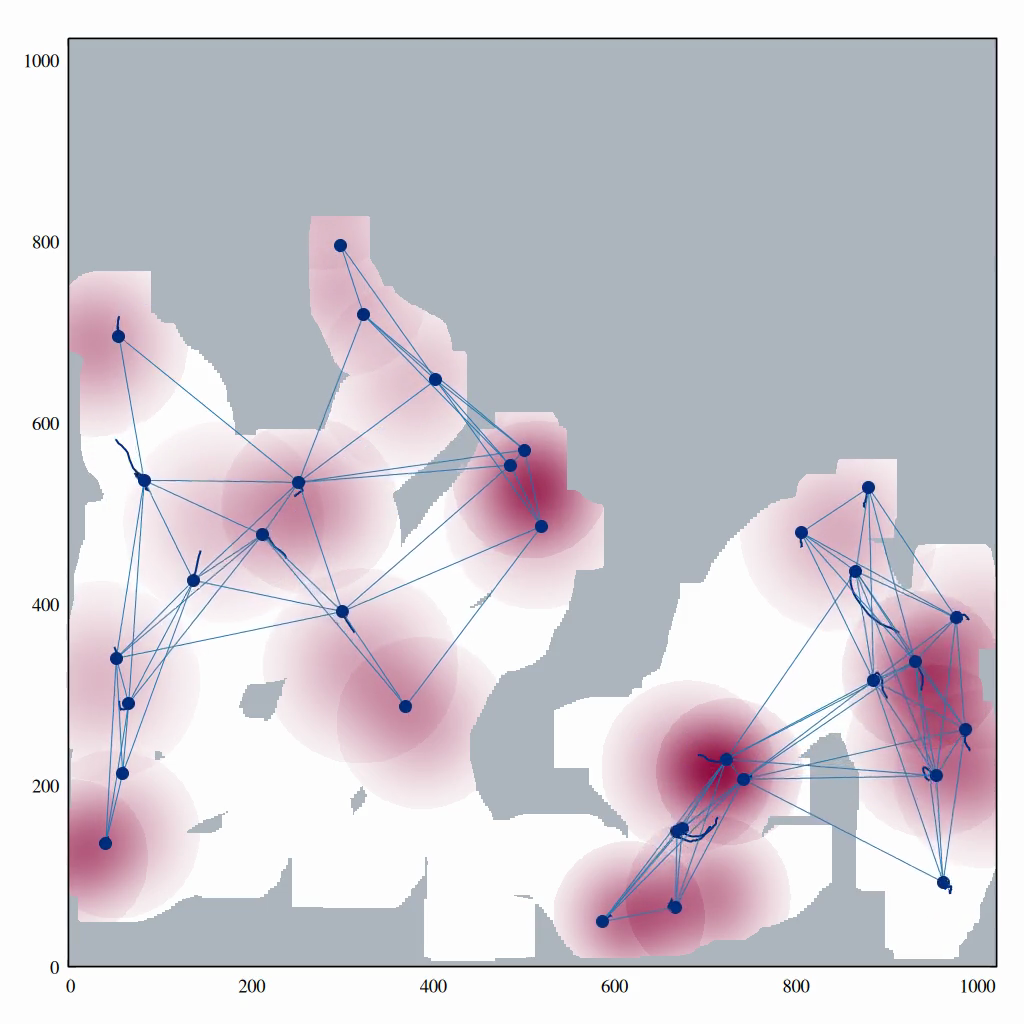}}\hspace{0.1cm}
    \subfloat{\includegraphics[width=0.23\textwidth, trim={70px 59px 29px 40px}, clip]{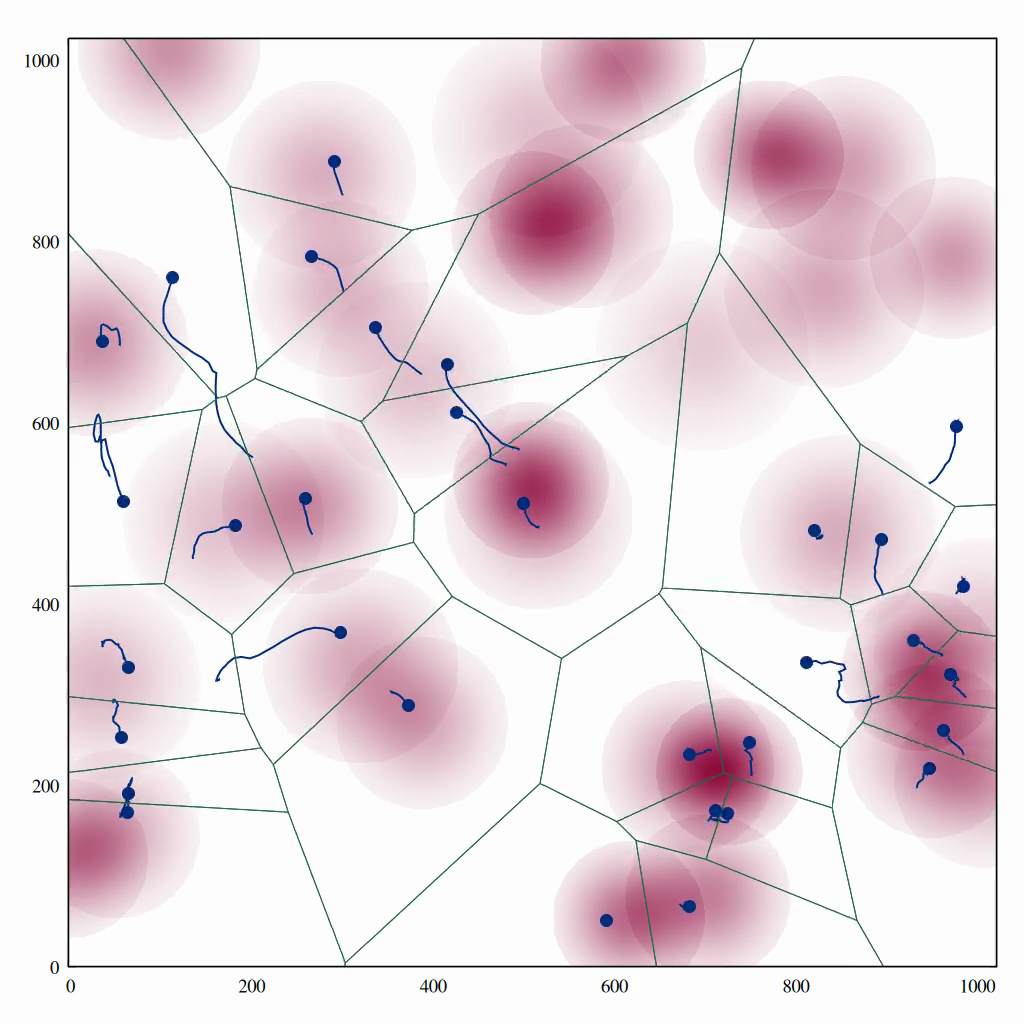}}\hspace{0.1cm}
    \caption{
        We experiment with different rules for generating the initial positions of the robots (blue dots).
        In each case, we define a subset of the environment over which we uniformly sample the positions.
        We explore restricting the launch positions to a small square placed on the outskirts of the environment (\textbf{Top}) and a band located near the bottom edge (\textbf{Bottom}).
        The areas of the map that have not been seen by any robot are colored gray and communication between robots is represented by a blue edge between nodes.
        From left to right, snapshots of the experiments are at $0$, $200$, $400$ and $600$ timesteps. 
        The complete IDF and induced Voronoi tessellation \eqref{eq:voronoi_partitioning} are shown at step $600$.
    }
    \label{fig:init-configurations}
\end{figure*}
Thus far, we have evaluated the policies on synthetic importance density functions. 
We now test performance on IDFs informed by real-world traffic light locations. 
These maps are constructed by capturing $(\SI{1024}{\meter} \times \SI{1024}{\meter})$ neighborhoods from 50 U.S. cities and placing a Gaussian at each traffic light.

We roll out MADP, LPAC-K3, and DCVT on each city for $600$ timesteps and record the mean normalized coverage cost (\fgref{fig:coverage-real-world}).
MADP achieves the lowest cost in $32$ out of $50$ cities, indicated by the red markers denoting the best policy in each case. 
Table \ref{tab:realworldscenarios} reports the mean cost $\pm$ standard error across all $50$ cities, reaffirming MADP’s superior performance in out-of-distribution conditions.

\begin{table}[t] \centering
    \caption{Mean Normalized Coverage Cost of Real-World Configurations}
    \begin{tabular}{lccc}
      \toprule
      Scenario & DCVT & LPAC-K3 & MADP \\
      \midrule
      50 Cities   & $0.63 \pm 0.03$ & $0.45 \pm 0.03$ & $\mathbf{0.39} \pm 0.03$ \\
      \midrule
      Uniform  & $0.72\pm0.01$     & $0.48\pm0.01$    & $\mathbf{0.46}\pm0.01$              \\
      Square   & $0.71\pm0.02$     & $0.46\pm0.02$     & $\mathbf{0.44}\pm0.02$     \\
      Line     & $0.76\pm0.02$     & $0.20\pm0.01$    & $\mathbf{0.17}\pm0.01$      \\
      \bottomrule
    \end{tabular}
    \label{tab:realworldscenarios}
\end{table}

To further assess adaptability to real-world deployments, we evaluate three initialization scenarios for robot positions (Table \ref{tab:realworldscenarios}):
(i) Uniform, (ii) Square, and (iii) Line; \fgref{fig:init-configurations}.
In all cases, the positions of each robot $i$, $(x, y)_{i}$, are sampled from the uniform distributions $x \sim \mathcal{U}(x_{\min},x_{\max})$ and $y \sim \mathcal{U}(y_{\min},y_{\max})$.
The difference is in the range of values $x$ and $y$ can take during the initialization of the robots in the environment.
In (i), robots can be initialized anywhere in the environment, $x, y \in [0, 1024]\si{\meter}$.
This is the same configuration used when generating the dataset.
We implement (ii) by uniformly sampling within a smaller square $(\SI{102}{\meter} \times \SI{102}{\meter})$ centered at coordinate $(\SI{166.25}{\meter}, \SI{166.25}{\meter})$ in the lower left corner, simulating, e.g., the deployment of quadrotors from a launch pad.
Thus for the (ii), we have,  $x, y \in [115.25, 217.25]\si{\meter}$.
For (iii), we define a tighter bound on $y$ only, creating a band of width $\SI{256}{\meter}$ for the initial positions.
In this case, we have that $x \in [0, 1024]\si{\meter}$ and $y \in [96, 352]\si{\meter}$.
For each starting configuration, we consider 50 unseen environments and $600$ timesteps.
In all cases, MADP outperforms the baselines, underscoring the robustness of the policy to initial conditions and its capacity to effectively explore the environment.

\subsection{Scalability and Transferability}
\begin{figure*}
    \centering
    \subfloat{\includegraphics[width=0.45\textwidth]{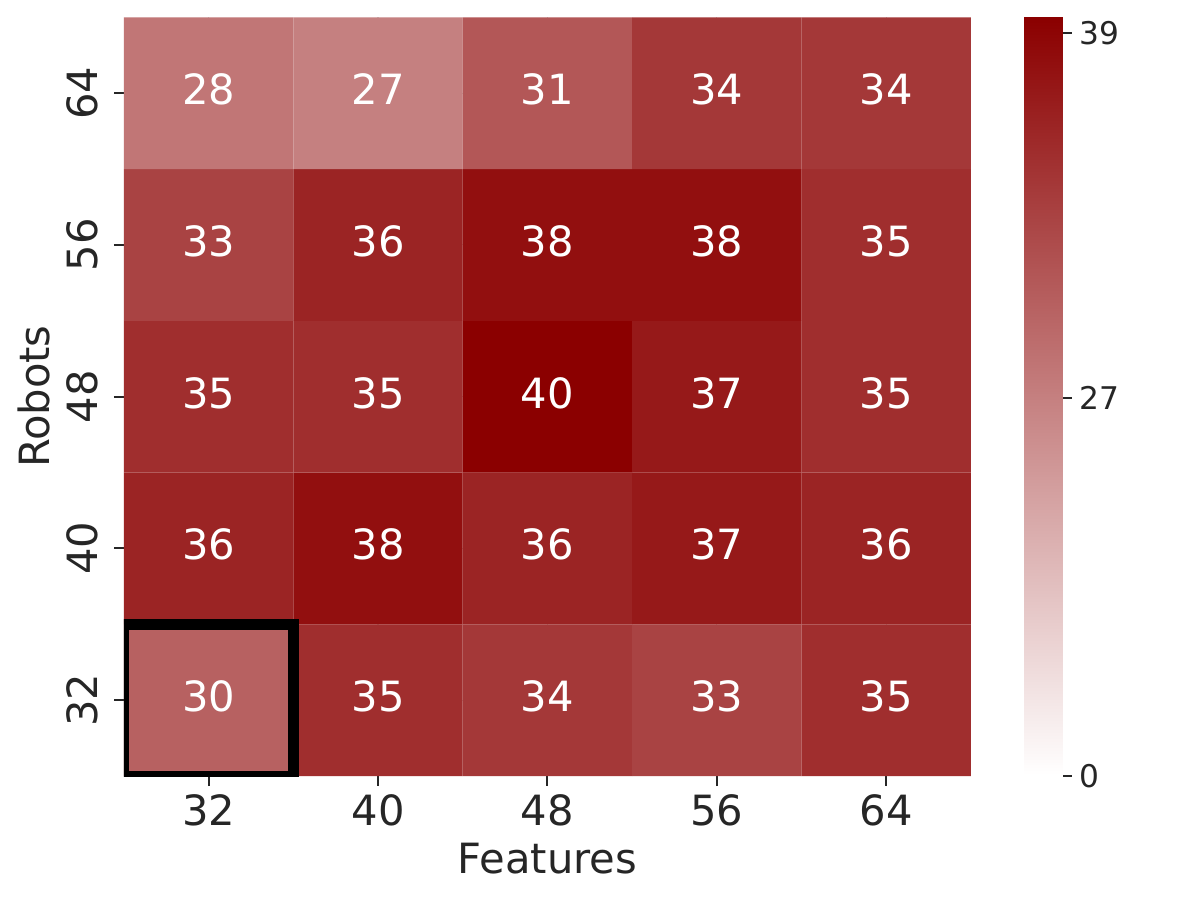}}\hspace{0.1cm}
    \subfloat{\includegraphics[width=0.45\textwidth]{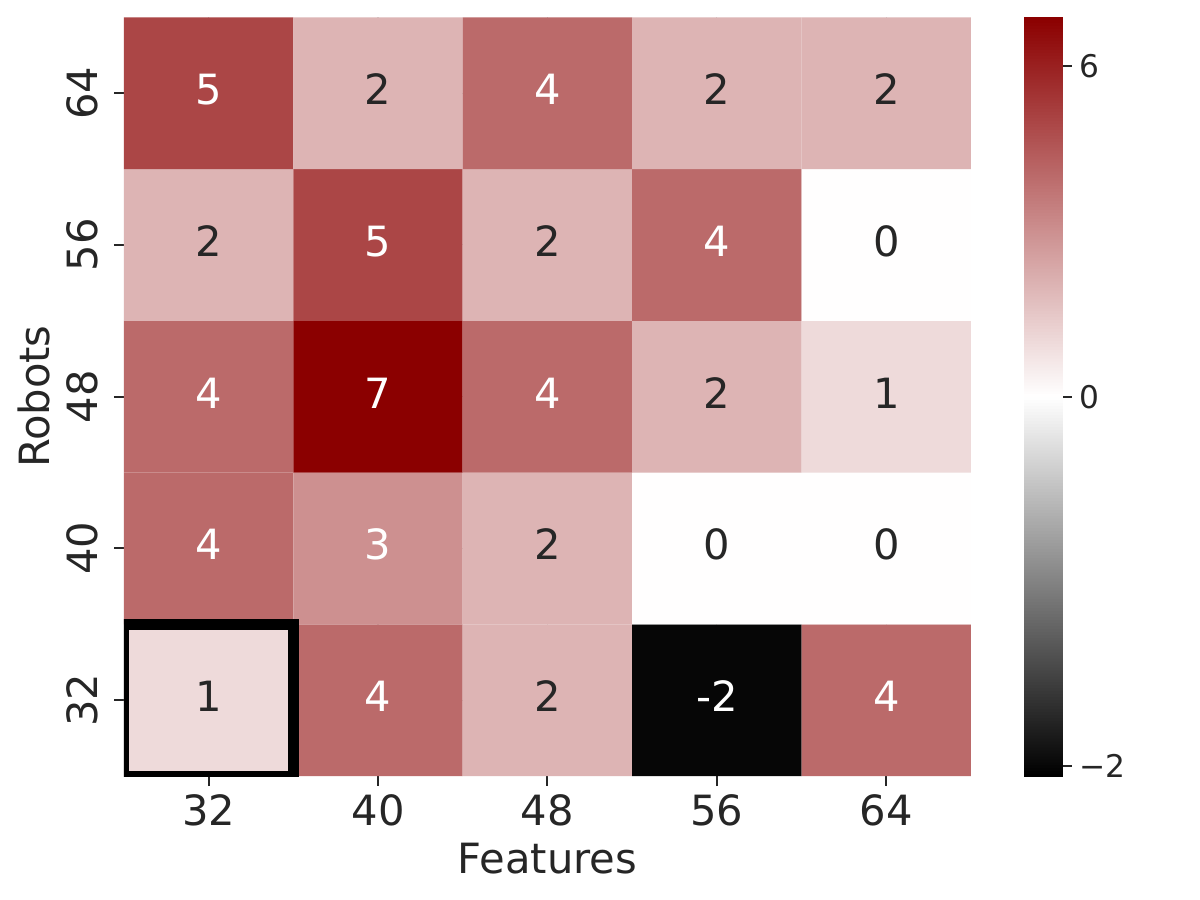}}\hspace{0.1cm}
    \caption{
    Each heatmap is a comparison against a baseline: against DCVT (\textbf{Left}) and against LPAC-K3 (\textbf{Right}).
    Each cell in the heatmap shows the percentage difference for evaluation of MADP against the respective baseline on environments with varying numbers of robots ($N$) and features ($F$). A positive value indicates an improvement over the baseline.
    MADP and LPAC-K3 are both trained on $N = F = 32$.
    }
    \label{fig:scalability}
\end{figure*}
As a final evaluation, we study the scalability and transferability of the learned policy, originally trained with $N=32$ robots and $F=32$ features.
\fgref{fig:scalability} reports the relative improvement of MADP over baselines DCVT and LPAC-K3, measured as $(\overline{\text{Baseline}} - \overline{\text{MADP}}) / \overline{\text{Baseline}}$, where $\overline{\cdot}$ denotes the average over trials.

For each setting we test the algorithms in 50 unseen environments. While performance decreases when the number of robots is reduced, MADP shows clear gains as the swarm size scales, outperforming the baseline in larger teams. Sensitivity to the number of features $F$ is minimal, indicating strong out-of-distribution generalization. Overall, these results highlight the scalability and transferability of MADP to settings with larger robot populations.

\section{Conclusion}
\label{sec:conclusion}
We propose MADP, a scalable, diffusion-based control policy for decentralized multi agent systems.
The novelty of MADP is in the combination of generative diffusion models with spatial transformers to enable sampling from the distribution of navigation policies for large-scale teams.
We demonstrate MADP on the coverage control task, showing that it exceeds the performance of the state-of-the-art LPAC-K3 across several variations of the problem.
MADP excels when areas of interest are smaller, which makes the problem more challenging.
We posit that this could be the result of the stochasticity of GDMs enhancing exploration through perturbation.
The diversity of trajectories produced by MADP motivates further investigation; we plan to explore ways to exploit this property in future work.
Guidance control and model predictive path integral (MPPI) control are promising approaches that leverage the use of these diverse stochastic trajectories to further improve performance.

\appendix
\section{Baselines}\label{app:baselines}
The coverage control problem can be reduced to finding centroidal Voronoi tessellations (CVT) of the environment.
A cost function is minimized by assigning each robot to a region of the space and performing gradient descent toward a local minimum.
A well-known algorithm for this problem is that proposed by Lloyd~\cite{lloyd1982}.
This original version assumes full access to the state of the environment.
In decentralized systems, robots exchange each other's positions and local sensing information, as proposed in~\cite{cortes2002}.

We consider several baselines based on finding Voronoi partitions, with different levels of observability.
The relevant benchmarks are summarized as follows:

\begin{itemize}
    \item \textbf{Clairvoyant:} Centralized algorithm with full access to the robot positions and the entire IDF.
    \item \textbf{Decentralized CVT (DCVT):} Decentralized algorithm with information exchange limited by the communication range.
      Each robot has access to the positions of its neighbors.
      Knowledge about the IDF is limited to local, historical sensory data.
    \item \textbf{LPAC-K3 \cite{agarwal_lpac_2025}:} State-of-the-art, learning-based benchmark. The algorithm is decentralized with information exchange limited by the communication range, each robot having access to the positions of the neighbors, and the sensing is local.
\end{itemize}




\bibliographystyle{IEEEtran}
\bibliography{IEEEabrv,MADP-zotero, MADP-manual}  

\begin{thebibliography}{10}
\providecommand{\url}[1]{#1}
\csname url@samestyle\endcsname
\providecommand{\newblock}{\relax}
\providecommand{\bibinfo}[2]{#2}
\providecommand{\BIBentrySTDinterwordspacing}{\spaceskip=0pt\relax}
\providecommand{\BIBentryALTinterwordstretchfactor}{4}
\providecommand{\BIBentryALTinterwordspacing}{\spaceskip=\fontdimen2\font plus
\BIBentryALTinterwordstretchfactor\fontdimen3\font minus \fontdimen4\font\relax}
\providecommand{\BIBforeignlanguage}[2]{{%
\expandafter\ifx\csname l@#1\endcsname\relax
\typeout{** WARNING: IEEEtran.bst: No hyphenation pattern has been}%
\typeout{** loaded for the language `#1'. Using the pattern for}%
\typeout{** the default language instead.}%
\else
\language=\csname l@#1\endcsname
\fi
#2}}
\providecommand{\BIBdecl}{\relax}
\BIBdecl
\renewcommand{\BIBentryALTinterwordstretchfactor}{4}

\bibitem{agarwal_lpac_2025}
S.~Agarwal, R.~Muthukrishnan, W.~Gosrich, V.~Kumar, and A.~Ribeiro, ``\href{https://doi.org/10.1109/TRO.2025.3619047}{{LPAC}: Learnable perception-action-communication loops with applications to coverage control},'' \emph{{IEEE} Trans. Robot.}, vol.~41, pp. 5986--6005, 2025.

\bibitem{cortes2002}
J.~Cort\'es, S.~Mart\`inez, T.~Karata\c{s}, and F.~Bullo, ``\href{https://doi.org/10.1109/TRA.2004.824698}{Coverage control for mobile sensing networks},'' \emph{{IEEE} Trans. Robot. Autom.}, vol.~20, no.~2, pp. 243--255, 2004.

\bibitem{longrl}
P.~Long, T.~Fan, X.~Liao, W.~Liu, H.~Zhang, and J.~Pan, ``\href{https://doi.org/10.1109/ICRA.2018.8461113}{Towards optimally decentralized multi-robot collision avoidance via deep reinforcement learning},'' in \emph{Proc. IEEE Int. Conf. Robot. Automat.}, 2018, pp. 6252--6259.

\bibitem{lignn}
Q.~Li, F.~Gama, A.~Ribeiro, and A.~Prorok, ``\href{https://doi.org/10.1109/IROS45743.2020.9341668}{Graph neural networks for decentralized multi-robot path planning},'' in \emph{Proc. IEEE/RSJ Int. Conf. Intell. Robots Syst.}, 2020, pp. 11\,785--11\,792.

\bibitem{ho_ddpm_2020_nips}
J.~Ho, A.~Jain, and P.~Abbeel, ``Denoising diffusion probabilistic models,'' in \emph{Proc. Int. Conf. Neural Inf. Process. Syst.}, ser. NIPS '20.\hskip 1em plus 0.5em minus 0.4em\relax Red Hook, NY, USA: Curran Associates Inc., 2020.

\bibitem{chi2024diffusionpolicy}
C.~Chi \emph{et~al.}, ``\href{https://doi.org/10.1177/02783649241273668}{Diffusion policy: Visuomotor policy learning via action diffusion},'' \emph{Int. J. Robot. Res.}, vol.~44, no. 10-11, pp. 1684--1704, 2025.

\bibitem{janner_planning_2022}
M.~Janner, Y.~Du, J.~Tenenbaum, and S.~Levine, ``\BIBforeignlanguage{en}{\href{https://proceedings.mlr.press/v162/janner22a.html}{Planning with diffusion for flexible behavior synthesis}},'' in \emph{\BIBforeignlanguage{en}{Proc. Int. Conf. Machine Learning}}.\hskip 1em plus 0.5em minus 0.4em\relax PMLR, Jun. 2022, pp. 9902--9915, iSSN: 2640-3498.

\bibitem{luo2024potentialbaseddiffusionmotion}
Y.~Luo, C.~Sun, J.~B. Tenenbaum, and Y.~Du, ``\href{https://openreview.net/forum?id=Qb68Rs0p9f}{Potential based diffusion motion planning},'' in \emph{Proc. Int. Conf. Machine Learning}, 2024.

\bibitem{carvalho_motion_2024}
J.~Carvalho, A.~Le, M.~Baierl, D.~Koert, and J.~Peters, ``\href{https://doi.org/10.1109/IROS55552.2023.10342382}{Motion planning diffusion: learning and planning of robot motions with diffusion models},'' in \emph{Proc. IEEE/RSJ Int. Conf. Intell. Robots Syst.}, 2023.

\bibitem{jiang_motiondiffuser_2023}
C.~Jiang, A.~Cornman, C.~Park, B.~Sapp, Y.~Zhou, and D.~Anguelov, ``{MotionDiffuser}: Controllable multi-agent motion prediction using diffusion,'' in \emph{Proc. IEEE Conf. Computer Vision and Pattern Recognition}, June 2023, pp. 9644--9653.

\bibitem{zhu_madiff_2025}
Z.~Zhu \emph{et~al.}, ``\href{https://openreview.net/forum?id=PvoxbjcRPT}{{MAD}iff: offline multi-agent learning with diffusion models},'' in \emph{Proc. Int. Conf. Neural Inf. Process. Syst.}, 2024.

\bibitem{shaoul_multirobot_2025}
Y.~Shaoul, I.~Mishani, S.~Vats, J.~Li, and M.~Likhachev, ``Multi-robot motion planning with diffusion models,'' in \emph{The Thirteenth International Conference on Learning Representations (ICLR)}, 2025.

\bibitem{liang2025discreteguideddiffusionscalablesafe}
J.~Liang, S.~Koenig, and F.~Fioretto, ``Discrete-guided diffusion for scalable and safe multi-robot motion planning,'' \emph{The 40th Annual AAAI Conference on Artificial Intelligence}, 2026.

\bibitem{ding2025swarmdiffswarmrobotictrajectory}
K.~Ding \emph{et~al.}, ``Swarmdiff: Swarm robotic trajectory planning in cluttered environments via diffusion transformer,'' in \emph{Proceedings of the IEEE/CVF Conference on Computer Vision and Pattern Recognition (CVPR) Workshops}, June 2025, pp. 4203--4212.

\bibitem{liang2025simultaneousmultirobotmotionplanning}
\BIBentryALTinterwordspacing
J.~Liang, J.~K. Christopher, S.~Koenig, and F.~Fioretto, ``Simultaneous multi-robot motion planning with projected diffusion models,'' in \emph{Proceedings of the 42nd International Conference on Machine Learning}, ser. Proceedings of Machine Learning Research, A.~Singh \emph{et~al.}, Eds., vol. 267.\hskip 1em plus 0.5em minus 0.4em\relax PMLR, 13--19 Jul 2025, pp. 37\,162--37\,180. [Online]. Available: \url{https://proceedings.mlr.press/v267/liang25e.html}
\BIBentrySTDinterwordspacing

\bibitem{gosrich2021coveragecontrolmultirobotsystems}
W.~Gosrich \emph{et~al.}, ``\href{https://doi.org/10.1109/ICRA46639.2022.9811854}{Coverage control in multi-robot systems via graph neural networks},'' in \emph{Proc. IEEE Int. Conf. Robot. Automat.}, 2022, pp. 8787--8793.

\bibitem{song_denoising_2020_iclr}
J.~Song, C.~Meng, and S.~Ermon, ``\href{https://openreview.net/forum?id=St1giarCHLP}{Denoising diffusion implicit models},'' in \emph{Proc. Int. Conf. Learn. Representations}, 2021.

\bibitem{su2023roformerenhancedtransformerrotary}
\BIBentryALTinterwordspacing
J.~Su, M.~Ahmed, Y.~Lu, S.~Pan, W.~Bo, and Y.~Liu, ``{RoFormer}: Enhanced transformer with rotary position embedding,'' \emph{Neurocomput.}, vol. 568, no.~C, Feb. 2024. [Online]. Available: \url{https://doi.org/10.1016/j.neucom.2023.127063}
\BIBentrySTDinterwordspacing

\bibitem{owerko2025mast}
\BIBentryALTinterwordspacing
D.~Owerko, F.~Vatnsdal, S.~Agarwal, V.~Kumar, and A.~Ribeiro, ``{MAST}: Multi-agent spatial transformer for learning to collaborate,'' 2025. [Online]. Available: \url{https://arxiv.org/abs/2509.17195}
\BIBentrySTDinterwordspacing

\bibitem{vaswani_attention_2017}
A.~Vaswani \emph{et~al.}, ``\href{https://proceedings.neurips.cc/paper_files/paper/2017/file/3f5ee243547dee91fbd053c1c4a845aa-Paper.pdf}{Attention is all you need},'' in \emph{Advances in Neural Information Processing Systems}, I.~Guyon \emph{et~al.}, Eds., vol.~30.\hskip 1em plus 0.5em minus 0.4em\relax Curran Associates, Inc., 2017.

\bibitem{OpenStreetMap_contributors}
\BIBentryALTinterwordspacing
{OpenStreetMap contributors}, ``Openstreetmap,'' 2025, oDbL 1.0. See https://www.openstreetmap.org/copyright. [Online]. Available: \url{https://www.openstreetmap.org}
\BIBentrySTDinterwordspacing

\bibitem{lloyd1982}
S.~Lloyd, ``Least squares quantization in pcm,'' in \emph{Transactions on Information Theory}.\hskip 1em plus 0.5em minus 0.4em\relax IEEE, 1982, pp. 129--137.

\end{thebibliography}
\end{document}